\pdfoutput=1

\documentclass[11pt]{article}

\usepackage[]{acl}
\usepackage{times}
\usepackage{latexsym}
\usepackage{multirow}
\usepackage{booktabs}
\usepackage{amssymb}
\usepackage{xspace}
\usepackage{array}
\usepackage[scaled=0.8]{beramono} 
\usepackage{enumitem}
\usepackage{color, colortbl}
\usepackage[T1]{fontenc}

\usepackage[utf8]{inputenc}

\usepackage{amsmath}
\usepackage{amssymb,amsmath,amsthm,enumitem}
\usepackage{graphicx}
\usepackage{microtype}
\usepackage{longtable}
\usepackage{comment}
\usepackage{csquotes}
\usepackage{soul}

\newcommand{\aref}[1]{\hyperref[#1]{Appendix~\ref{#1}}}

\definecolor{darkergreen}{rgb}{0.0,0.4,0.0}

\newcommand{\Qq}[1]{\textbf{\textsc{#1?}}\xspace}
\newcommand{\Qrule}{\Qq{rule-based}}
\newcommand{\Qhate}{\Qq{toxic}}
\newcommand{\Qmatch}{\Qq{match}}
\newcommand{\Qpurpose}{\Qq{purpose}}
\newcommand{\Qgap}{\Qq{gap}}
\usepackage{subcaption}

\title{Toxicity Detection is NOT all you Need: \\Measuring the Gaps to Supporting Volunteer Content Moderators through a User-Centric Method}

\author{
    Yang Trista Cao\textsuperscript{\textnormal{1}}
    ~~~~Lovely-Frances Domingo\textsuperscript{\textnormal{1}}
    ~~~~Sarah Ann Gilbert\textsuperscript{\textnormal{3}}\\
  \textbf{Michelle L. Mazurek\textsuperscript{\textnormal{1}}}
  ~~~~\textbf{Katie Shilton\textsuperscript{\textnormal{1}}}
  ~~~~\textbf{Hal Daum\'e III\textsuperscript{\textnormal{1,2}}}
  \\
    $^1$University of Maryland, College Park \qquad $^2$Microsoft Research \qquad $^3$Cornell University
    \\
    \texttt{\{ycao95,lovely,mmazurek,kshilton,hal3\}@umd.edu}
    \\
    \texttt{sag284@cornell.edu}
  }

\begin{document}
\maketitle

\begin{abstract}
Extensive efforts in automated approaches for content moderation have been focused on developing models to identify toxic, offensive, and hateful content with the aim of lightening the load for moderators.
Yet, it remains uncertain whether improvements on those tasks have truly addressed moderators' needs in accomplishing their work.
In this paper, we surface gaps between past research efforts that have aimed to provide automation for aspects of content moderation and the needs of volunteer content moderators, regarding identifying violations of various moderation rules. 
To do so, we conduct a \textit{model review} on Hugging Face to reveal the availability of models to cover various moderation rules and guidelines from three exemplar forums. 
We further put state-of-the-art LLMs to the test, evaluating how well these models perform in flagging violations of platform rules from one particular forum.
Finally, we conduct a user survey study with volunteer moderators to gain insight into their perspectives on useful moderation models. 
Overall, we observe a non-trivial gap, as missing developed models and LLMs exhibit moderate to low performance on a significant portion of the rules. 
Moderators' reports provide guides for future work on developing moderation assistant models.  
\end{abstract}

\section{Introduction} \label{sec:sections/intro} Content moderation guards online forums against hostility and extremism while maintaining community norms, ensuring the forums remain healthy and open to all participants. While many platforms pay for this service, others, such as Reddit, Discord, Facebook, and Twitch, use a hybrid model, relying on the labor of volunteers. Yet, behind the screen, being a volunteer content moderator is time- and emotionally-draining work. Moderators frequently deal with abusive language, sensitive posts, and unpleasant interactions with users \cite{seering2019moderator, gilbert2020run, dosono2019moderation, wohn2019volunteer, jiang2019moderation}, often doing this work in addition to full-time jobs. To support these volunteers, efforts have been made to develop models, such as Google Perspective API\footnote{\url{https://perspectiveapi.com/}} and OpenAI undesired content detection~\cite{Markov_2023}, that can automatically identify content for removal in order to alleviate moderators' workload.

Although these systems have shown great success in detecting ``undesired'' content, they primarily focus on toxic content. Yet, content moderation encompasses more than toxicity detection,\footnote{We use ``toxicity detection'' as an umbrella term for hate speech detection, incivility detection, etc.} particularly in platforms that leverage volunteer moderation within smaller communities hosted by the site. 
For example, Reddit is a platform consisting of various communities, known as ``subreddits,'' focused on a diverse set of topics, and each subreddit has its own moderation rules. 
\citet{Fiesler_2018} conducted a study to explore various subreddit rules, consolidating similar ones, and arrived at $25$ distinct rule types. 
Hence, in order to support moderators in detecting potential rule-violating content, content moderation tools need to support much more than just toxicity detection. 

In this paper, we aim to assess to what extent current natural language processing (NLP) models can serve the wide spectrum of moderation rules so that they can be helpful in assisting moderators.
First, to understand the functions previous automated content moderation models have focused on, we conduct a \textit{model review} on Hugging Face (HF) with rules from three subreddits as exemplars.
This allows us to gauge the progress of past model developments in covering various moderation rules.
We use model review as opposed to the more common literature review to gain a technical understanding of the existing models' functions.
In addition to examining models that are built to handle specific tasks, we also assess so-called ``general-purposed'' large language models' (LLMs') capability in covering various moderation rules. 
We evaluate GPT-4 and Llama-2 on a new evaluation dataset that we collected, consisting of moderation decisions from \textit{r/AskHistorians} and covering a wide range of rules.
Finally, using the models' performance on this new dataset as an empirical grounding, we conducted a survey study with active moderators from \textit{r/AskHistorians} ($N=11$). Through this survey study, we aim to gain insights into model users' perspectives on the performance requirements for useful moderation models on different rules.

We find a substantial gap in both existing NLP models and LLMs' performance when to cover the diverse set of moderation rules that subreddits employ.
Our analysis shows the majority of moderation rules from the three subreddits ($\sim 80\%$) are unrelated to toxicity detection, with nearly $70\%$ lacking an huggingface model designed for their resolution.
While one might hope that general-purpose LLMs could fill this gap, our experiments with GPT-4 and Llama-2 show that LLMs fall short in their ability to detect violations of many rules. 
Specifically, both GPT-4 and Llama-2 exhibited moderate to low precision and/or recall ($<70\%$) for half of the rules from \textit{r/AskHistorians}. 
Findings from our survey study also indicate that neither LLM has good enough performance to be useful for $6$ of $23$ \textit{r/AskHistorians} rules ($26\%$), including rules such as \texttt{Scope}, \texttt{Digression}, and \texttt{Sources}\footnote{For explanation of the rules, please refer to \autoref{tab:rules_ask} in appendix.}. 
Meanwhile, our survey study also shows that moderators are excited about an assistant model---they are okay with even imperfect tools if they are well-informed about their limitations.
For different kinds of rules, moderators have complex needs in terms of model precision and recall. For instance, they need high recall for complex rules (e.g. \texttt{Plagiarism} and \texttt{Digression}) and high precision for simple rules (e.g. \texttt{Current Event} and \texttt{Jokes and Humour}). 
Our study highlights the necessity of, and provides a guide for, future content moderation work to expand its focus beyond toxicity detection, encompassing a broader spectrum of moderation rules to meet the needs of moderators seeking automation assistance.

\section{Background and Related Work} \label{sec:sections/background} 
Volunteer moderators have been a staple of online content regulation from the earliest days of the internet, tackling issues such as spam \cite{seering2020reconsidering}, trolling \cite{binns2012don}, and abuse \cite{dibbell1994rape}. Currently, volunteers contribute moderation efforts on nearly all major platforms. For example, volunteers are responsible for moderation within Facebook's Groups, Discord's servers, Twitch's streams, Reddit's subreddits, and through X's (formally known as Twitter) Community Notes. These efforts bring significant value to platforms, with one study estimating that, at the baseline, volunteer moderators on Reddit collectively work 466 hours a day, amounting to a value of about 3.4 million USD per year \cite{li2022measuring}. 
With the workload and continuous exposure to online harassment \cite{gilbert2020run, wohn2019volunteer, dosono2019moderation} and toxic content, such as hateful language and violent or upsetting imagery, this moderation work can easily result in burnout.

A common way to mitigate burnout is through the development of tools that support moderation labor, particularly through automation. For example, prior study has found that automation can help at scale, supporting moderators of large communities with high levels of activity  \cite{kiene2020uses, seering2018social}. Automation is also helpful when communities experience unprecedented or unexpected growth, such in cases where communities receive an influx of new users \cite{kiene2016surviving} and in cases where communities are subject to sustained brigades of bad actors spamming hateful content \cite{han2023hate}. Mostly, automated tools are developed and maintained by moderators themselves. For example, the most commonly used tool on Reddit, Automod, was originally designed by a moderator before the company took over responsibility for its development and maintenance \cite{jhaver2019human} and moderators on Twitch have developed bots on the fly in response to hate raids \cite{han2023hate}. 
While automation can help support moderation work, it may also add to it in different ways—like needing to build and maintain a bot—and requires skill sets not all moderators have \cite{jhaver2019human}.

Meanwhile, the NLP field has a long line of work on developing automated models to detect toxic content, which can be used to support moderation work \citep[e.g.][]{jigsaw_2019,Pavlopoulos_2020, park-fung-2017-one,zampieri-etal-2019-predicting}.
Beyond that, \citet{multilingual_perspective} extended beyond English content and proposed the multilingual Perspective API to detect offensive and hateful content from a diverse range of languages.
Moreover, some study delves into various types of toxic content, trying to improve the nuance of detection models \cite{Markov_2023, Price_2020}.
However, these studies primarily focus on detecting toxic content, which is merely a subset of moderation work. Therefore, our study uses Reddit rules as an example to identify gaps in NLP moderation models within a wide spectrum of rules.

Recently, \citet{Kumar_2023} tested LLMs' performance on rule-based moderation with rules from $95$ subreddits. They evaluated GPT models (3, 3.5, 4) on a dataset consisting of removed Reddit posts as violating and not violating content. They conclude that the GPT models are effective in conducting moderation on subreddits like \textit{r/Movies} but struggle with other subreddits such as \textit{r/askscience} and \textit{r/AskHistorians}.
Our study builds on previous results by studying gaps in both purpose-built moderation models and LLMs. Rather than focusing on the broad binary questions of removing or keeping posts, we focus on coverage for specific moderation rules. In addition, we include the perspectives of human moderators to understand how existing models address the needs of real-world moderation.
\section{Hugging Face Model Review} \label{sec:sections/model_review} \begin{figure*}[t]
    \centering
\includegraphics[width=1.8\columnwidth,angle=0,clip=true,trim=0 190 0 10]{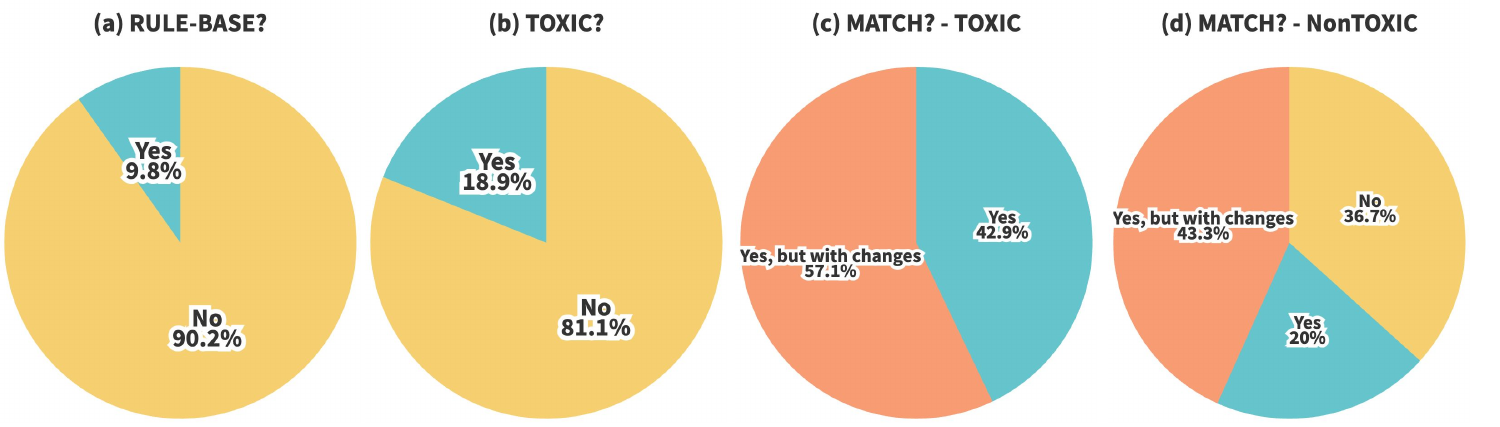}
    \caption{Model review annotation results. Figures (a) and (b) are the results of annotation questions \Qrule and \Qhate, respectively. Figures (c) and (d) are the results of question \Qmatch for rules related to toxicity detection and not related, respectively.}
    \label{fig:HF-results}
\end{figure*}

One traditional approach to understanding a scientific context is to perform a literature review. 
Because we are particularly interested in deployable models, we opt for a \textit{model review}, where instead of reviewing papers, we review publicly available models that can be adapted to assist content moderators in making moderation decisions, with the focus on Reddit rules.
We conduct the model review via Hugging Face (HF)\footnote{\url{https://huggingface.co/models}}, an open-source platform that provides pre-trained models for various NLP tasks, to understand the progress of model development from existing research in covering various moderation rules.
To conduct the model review, we gather rules from three subreddits and manually investigate for each rule whether there exists an HF model suitable for detecting such rule violations. 

\subsection{Method}
\citet{Chandrasekharan_2018} clustered 100 subreddits into six Meso groups plus four Micro groups based on similarity of the removed posts. 
Since Micro groups are subreddits, whose rules are more specific to the particular subreddit, we focus on the Meso groups, whose rules are shared across multiple subreddits within the group. We pick three subreddits -- \textit{r/Atheism, r/Movies, r/AskHistorian}, from three of the major Meso subreddit clusters and use their rules as representatives for the study. 
We verify the coverage of the three subreddits' rules by cross-referencing them with the list of $25$ rule types from \citet{Fiesler_2018}, ensuring that all text-based rule types are covered by at least one of the rules from the three subreddits. See appendix \autoref{tab:rule_types} for our list of rules and its cross-referencing with the list of rule types.

\paragraph{Procedure}
For each rule, we first crawl its rule description from Reddit and answer the following two questions based on the description. 

\noindent~~$\blacktriangleright$~\Qrule: Can detection of violations on this rule be done by rule-based approaches, e.g. by regular expressions (yes or no)? We first mark out the rules that can easily be handled by rule-based approaches, such as detecting reposts. 
If the answer is yes, we do not consider this rule any further; all other rules are fully 
analyzed using following steps. 

\noindent~~$\blacktriangleright$~\Qhate: Is this rule covered by toxicity detection (yes or no)? To understand what portion of the policies are related to toxicity detection, we mark the rules that can be handled by toxicity detectors, such as detecting harassment. 

Then, we search on HF for a \textit{matching model} to each rule. Based on the rule descriptions and our knowledge of previous NLP task names, we identify keywords for model searching. For example, for the rule \texttt{Personal Attacks or Flaming}: \textit{``Keep things civil. Avoid fighting words and personal attacks. (applies to all forms of user content, including the user's name.)''}, we use keywords \texttt{doxxing} and \texttt{cyberbullying}.

We then search on HF by matching keywords with model names or model cards to find the most relevant model to each moderation rule. 
We traverse the top 20 search results to find the most relevant model, which we call the \emph{matching model}. If more than one model is relevant, we select the model with the highest number of likes on HF. If nothing relevant is found, we declare there is no matching model for this moderation rule. We also skip any model that does not contain a model card (i.e., no model description). See Tables \ref{tab:annotation_atheism}, \ref{tab:annotation_movies}, and \ref{tab:annotation_ask} in the appendix for the keywords we used and the matching models for each rule.


With the matching model, we then answer the following questions for each moderation rule:

\noindent~~$\blacktriangleright$~\Qmatch: Is there an HF model that can be used to detect violations of this rule? Possible answers are \textit{yes}; \textit{yes, but with changes}; or \textit{no}. We mark \textit{yes, but with changes} when the model is intended to be used for tasks similar to detecting violations of this rule, but some adjustment are needed for the model to be useful in this case. The adjustments needed are noted in the last question.

\noindent~~$\blacktriangleright$~\Qpurpose: What is the original purpose of this matching model? We note the intended use case for the model as claimed in the model card.

\noindent~~$\blacktriangleright$~\Qgap: In order to adapt this model to detect violations of this rule, what needs to be adjusted? We write the changes, if needed, for the model to be used for this rule, such as domain adaptation.

\subsection{Model Review Results}
As shown in \autoref{fig:HF-results}, among the $41$ moderation rules, $37$ ($90\%$) of the rules cannot be handled by rule-based approaches. Among these, only $7$ ($19\%$) are toxicity-related. This reemphasizes that models solely designed for detecting toxicity fall short in meeting the practical needs of content moderators.

Among the $7$ toxicity-related rules, we see that $4$ ($57\%$) require model modifications. Most of these modifications are customization for the rule, which we will describe later.
Even toxicity-related rules, despite having many developed models, do not have perfectly matching models. 

Among the $37$ non-toxicity-related rules, only $6$ ($20\%$) rules have matching HF models that can be applied directly. To see some evaluation results of these matching models, please refer to  \autoref{sec:matching_eval} in appendix. $13$ ($43\%$) of the rules require some model modifications in order for the matching model to be adapted for the rule, whereas $11$ ($37\%$) rules have no matching model at all. These rules include, for example, \texttt{low-effort posts} and \texttt{proselytizing} from \textit{r/Atheism}, and \texttt{no homework} and \texttt{no ``soapboxing'' or loaded questions} from \textit{r/AskHistorians}.

Overall, many rules lack a matching model. For the rules that have one, most require non-trivial model modifications to be covered.   
These results indicate a substantial gap between the functionalities of previously developed models and those needed for Reddit moderation.

For the \Qgap question, we primarily identify four types of gaps.
\begin{enumerate}[topsep=0pt,itemsep=-1ex,partopsep=1ex,parsep=1ex]
    \item The most common adjustment is domain adaptation to the rule-related topic or to Reddit texts. For example, the \texttt{No Ambiguous/Misleading/Inaccurate Information} rule in \textit{r/Movies} can be handled with a misinformation detection model, but the model needs to be adapted to title-like texts and information about movie releases. 
    \item Some models are developed with a limited scope of training data and thus may require extended training to be used for moderation. For instance, we found a plagiarism model that was trained on the Machine Paraphrase Corpus~\cite{mpc} consisting of $\sim 200k$ examples of original and paraphrases using two online paraphrasing tools. The plagiarism cases on Reddit may be more complicated than paraphrasing, and the potential plagiarism source is larger. Hence, the model needs an extension of scope in order to be useful.
    \item Models for some rules, especially toxicity-related rules, require customization. For example, \textit{r/Atheism} has the rule \texttt{Harassment or Bigotry} that prohibits harassment but permits curse words. A toxicity detection model can be used here but needs modification. 
    \item Some models can be applied to certain rules, but with a degree of stretching. 
For instance, violations of the rule \texttt{Off Topic} can potentially be caught by a topic classifier model, but the model requires some major changes to adapt to the subreddit and its related topics. 
\end{enumerate}

To see detailed annotations for each question, please refer to Tables \ref{tab:annotation_atheism}, \ref{tab:annotation_movies}, and \ref{tab:annotation_ask} in appendix.

\section{LLM Performance on Subreddit Rules} \label{sec:sections/model_test} \begin{figure*}[t]
    \centering
\includegraphics[width=1.5\columnwidth, clip=true,trim=0 10 0 10]{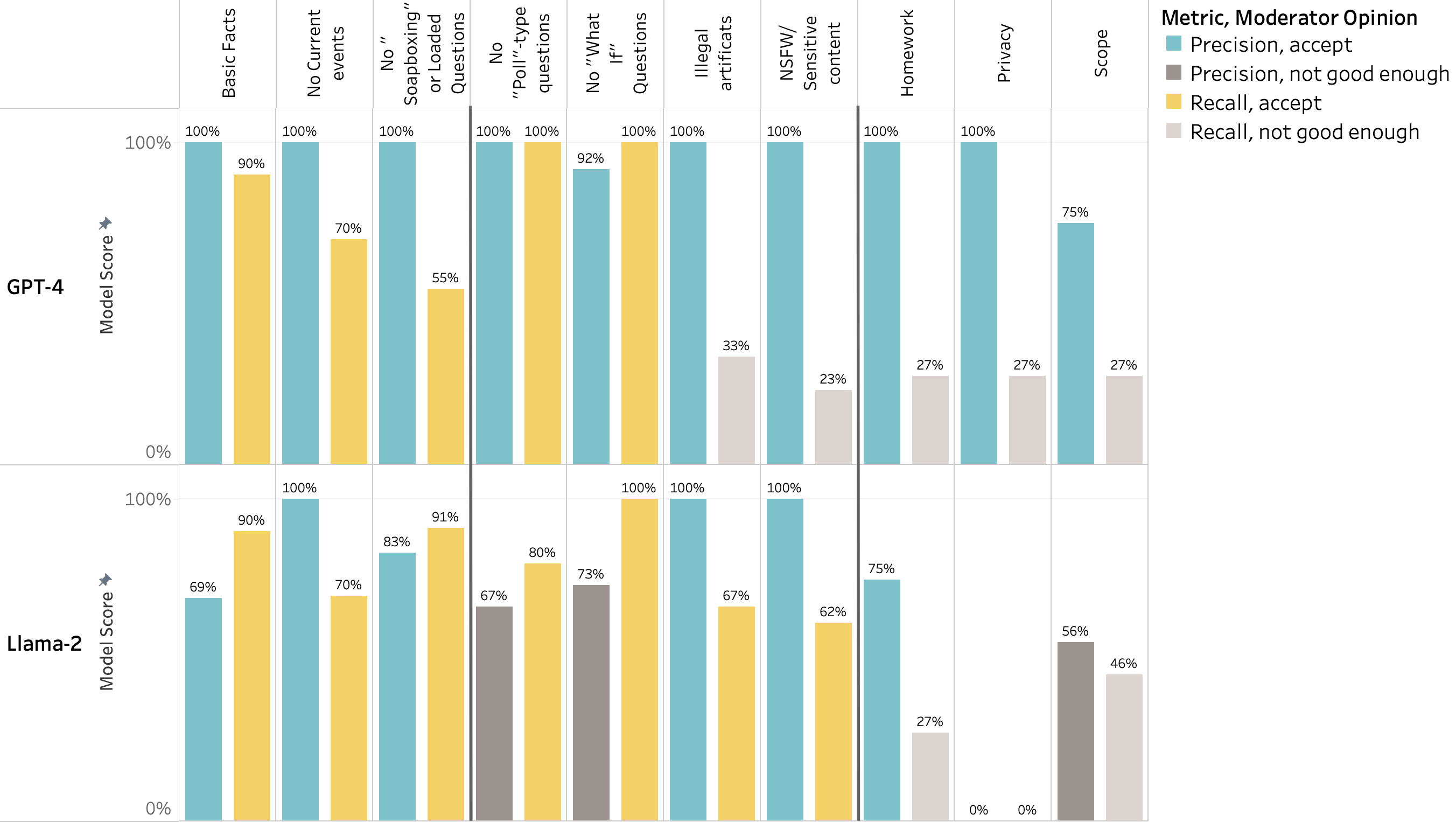}
    \caption{Model performance on detecting question posts that violate moderation rules with GPT-4 (top) and Llama-2 (bottom) models. The x-axis is the moderation rules for question posts. Each bar pair is the precision (left) and recall (right) scores on the specific rule. The ones marked grey are the model scores that moderators would not consider useful as a moderation assistant tool. The left-most rules have good performance from both of the LLMs; the rules in the middle have good performance from at least one of the LLMs; the right-most rules do not have good enough performance from either of the LLMs.}
    \label{fig:model-question}
\end{figure*}

\begin{figure*}[t]
    \centering
\includegraphics[width=1.5\columnwidth,clip=true,trim=0 10 0 10]{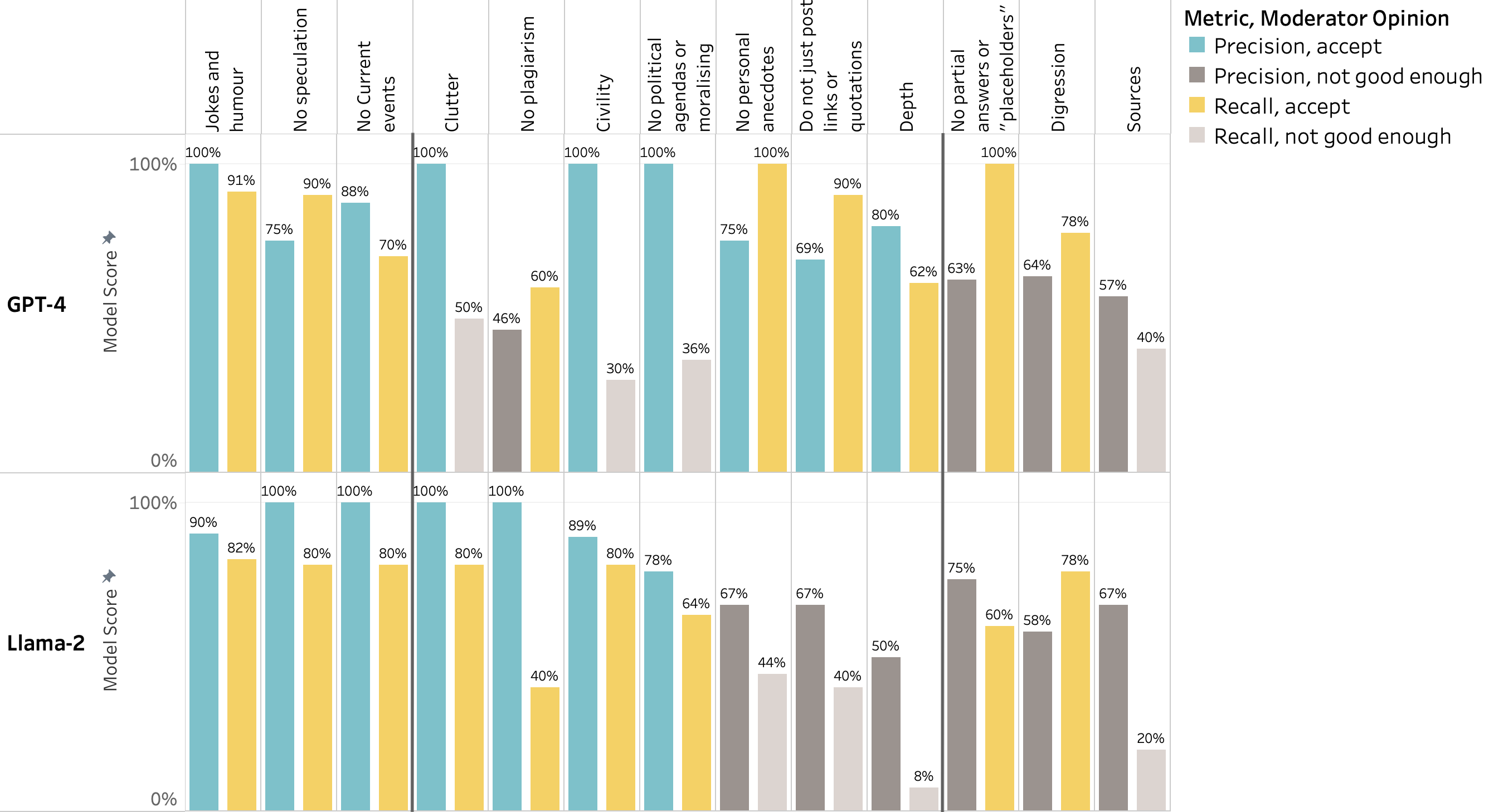}
    \caption{Model performance on detecting comment posts that violate moderation rules with GPT-4 (top) and Llama-2 (bottom) models. The x-axis is the moderation rules for comment posts. Each bar pair is the precision (left) and recall (right) scores on the specific rule. The ones marked grey are the model scores that moderators would not consider useful as a moderation assistant tool. The left-most rules have good performance from both of the LLMs; the rules in the middle have good performance from at least one of the LLMs; the right-most rules do not have good enough performance from either of the LLMs.}
    \label{fig:model-comment}
\end{figure*}


Finding a suitable NLP model for handling a moderation rule, let alone executing the necessary modifications to the model, is not trivial, and some rules simply lack a matching model to address their specific issues. It requires familiarity with NLP techniques and tasks, which may not be within the scope of a content moderator's abilities.
Question-answering-based LLM applications, on the other hand, may be more practical for content moderators to employ in assisting their jobs. 
Therefore, we also evaluate LLMs' performance in catching rule violations.
For this evaluation, we focus on the \textit{r/AskHistorians} subreddit, which has $23$ moderation rules, and two LLMs --- Llama-2 and GPT-4.

\subsection{Method}
\paragraph{Evaluation Dataset}
The evaluation dataset was created and annotated by a volunteer moderator for \textit{r/AskHistorians} who has expertise in both moderation and qualitative analysis.
For each rule, $11$--$13$ violating posts or comments were selected, save for posts violating the \texttt{Illegal artifacts} rule, which, due to the rareness of violations, was only represented $3$ times in the dataset. Most content was selected from \textit{r/AskHistory}, a Reddit community that is thematically similar to \textit{r/AskHistorians} but with different moderation rules. Some additional content was taken directly from \textit{r/AskHistorians}. 
We exclude all borderline cases and only select posts and comments that clearly violate the rules. 
In total, the dataset includes $101$ questions and $134$ comments.\footnote{In \textit{r/AskHistorians}, posts are questions users have asked and comments are responses to questions. Comments included in the dataset are not necessarily responses to the questions included in the dataset.} Because content often violates more than one rule, questions and comments included in the dataset often reflect multiple violations; however, annotations represent the most relevant or severe violation. For example, a comment containing only a joke would be annotated as violating the rule prohibiting joke responses, even though it would also violate rules on comprehensiveness or depth. The dataset also included $11$ questions and $10$ comments that do not violate any \textit{r/AskHistorians} rules. We use these data as our evaluation dataset.\footnote{The dataset, along with a datasheet, are available under a MIT licence at: \url{https://github.com/TristaCao/into_inclusivecoref}.}

\paragraph{Model Testing and Prompts}
To assess the performance of LLMs, we test GPT-4 and Llama-2-13b models. We conduct a pilot experiment with a small number of randomly selected examples from our dataset to determine the prompt for model assessment. 
Since moderators are mostly unfamiliar with prompt tuning and have limited time to learn this skill, we aim to test the models without extensively experimenting different wordings or examples of the prompt, roughly as they might be used by moderators.

First, as suggested in the prompting guidelines \cite{prompt}, we find that using rule descriptions in the prompt is helpful for the model to detect violations of the rule. 
Hence, we crawl rule descriptions from \textit{r/askHistorians} rule explanation web page.
As the descriptions from the subreddit web page are long and include many constructive suggestions for post-writers, we manually select useful information for moderation to include in the prompt. See appendix \autoref{tab:rules_ask} for the list of the rules and their descriptions.

Moreover, we experiment with both zero-shot and few-shot settings. 
In the few-shot setting, we provide the model with three examples that violated the rule and three examples that did not for question contents. For comment contents, we provide one example for each due to the model's input length limit. 
We realize that few-shot examples are helpful for the Llama-2 model to better understand the task, whereas the GPT-4 model does not have significant improvement from incorporating the examples. Therefore, for the main experiment, we test GPT-4 with the zero-shot setting and Llama-2 with the few-shot setting, but we also report the full set of experiments in the appendix.

Finally, following the suggestions from \citet{prompt}, we set the model's role as a moderator assistant and put the main question at the beginning of the prompt. In the few-shot setting, we also repeat the main question at the end of the prompt. See \autoref{sec:prompt} for the exact prompts we used.

    


\subsection{Moderator Evaluation}
To understand volunteer moderators' perspectives on the relative performance of LLMs, we also conducted an evaluation survey with active moderators from \textit{r/AskHistorians}\footnote{Approved by our institutional IRB, \#1704882-9.}.
The survey aimed to capture 1) whether moderators would use the LLM to help with their moderation work on each rule, given the LLM's performance (precision and recall) identifying violations for this rule, and 2) for each rule, what kind of model performance is needed to effectively assist moderators.

To answer the first question, we asked participants to evaluate each model's performance. We showed participants the LLMs' precision and recall scores for each rule from our experiment and asked them to choose among: 1. \textit{performance for this rule is good enough that I would use the tool},  2. \textit{I would use this tool for this rule if it could catch more violations} (need higher recall), 3. \textit{I would use this tool for this rule if it could be correct more often} (need higher precision), or \textit{4. both measures would have to improve for me to use this tool.}

For the second question, we asked participants to cluster moderation rules based on how important it is for the model to have high precision (or high recall) for each rule, compared to other rules. Options included: \textit{most important, less important,} and \textit{would not use a model for this rule}. We also asked for participants' comments on using LLMs to identify violating posts in general. See appendix Figures \ref{fig:survey-rate} and \ref{fig:survey-bucket} for the exact questions we asked.

Overall, by invitation, we recruited $11$ out of $36$ total active moderators from \textit{r/AskHistorians}. We obtained their consent at the beginning of the survey. Our participants spent an average of $25$ minutes completing the survey. In return, participants received compensation in the amount of $\$25$ USD upon completion of the survey in full.

\begin{figure}[t]
    \centering
\includegraphics[width=1\columnwidth]{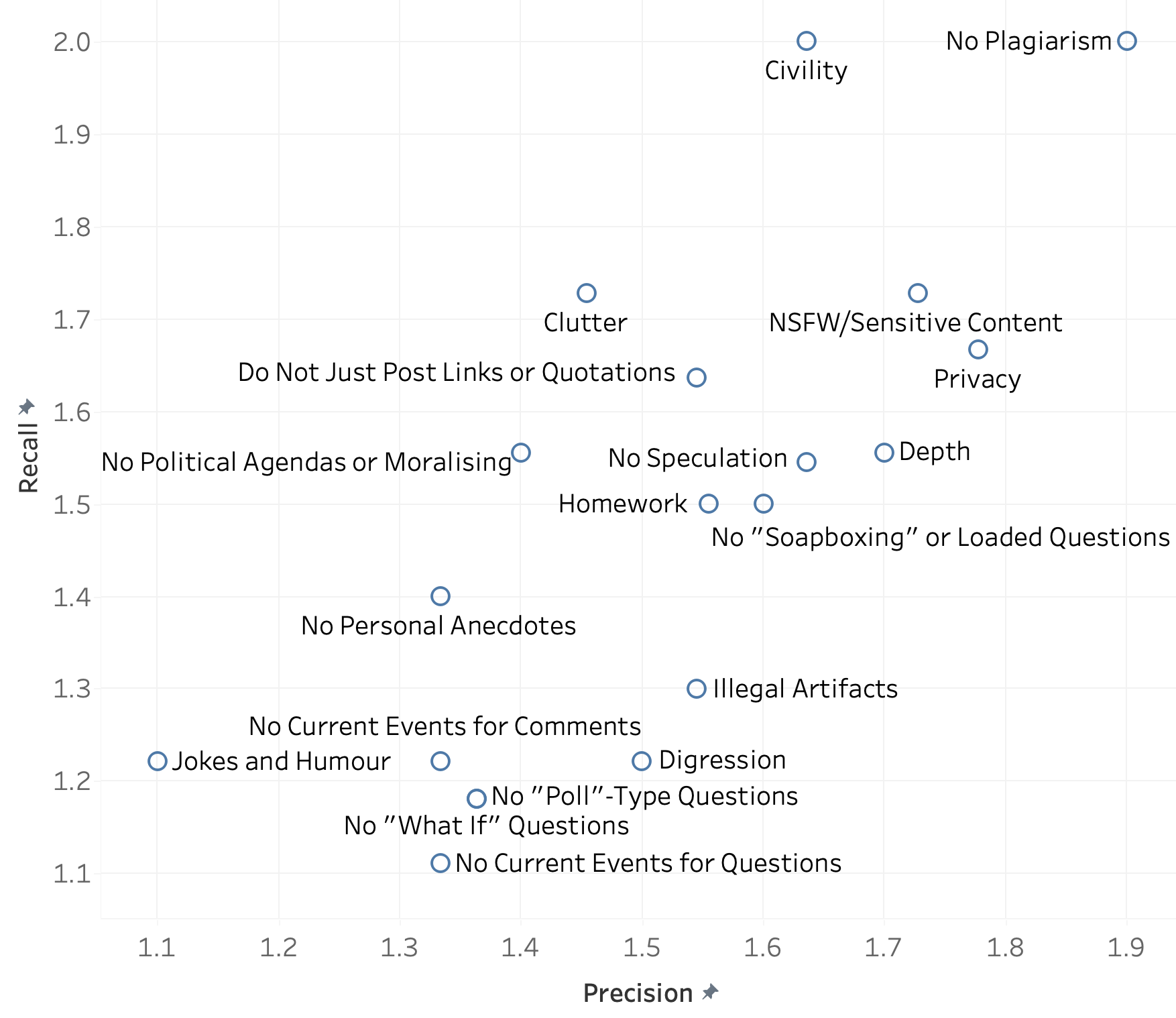}
    \caption{Clustering results from the survey study. Each point is a moderation rule from \textit{r/AskHistorians}. The x-axis shows the precision importance score, or how important it is for a model to have high precision for this rule. Similarly, the y-axis shows the recall importance score. Note that we removed rules for which at least three participants stated they would not use a model. }
    \label{fig:bucket}
\end{figure}

\subsection{Model Performance -- Results}
We tested GPT-4 and Llama-2 on the evaluation dataset as we described above. We use the first occurrence of ``Yes'' or ``No'' in the generated text as the model's output. 
The results are shown in \autoref{fig:model-question} and \autoref{fig:model-comment} (See appendix \autoref{fig:gpt-append} and \autoref{fig:llama-append} for the results of the models in both few-shot and zero-shot settings). 
The results are mixed -- some rules have both high precision and high recall from at least one model, while many of the rules have moderate to low precision, recall, or both. 
Neither of the two models is clearly better than the other in identifying violating contents for all the rules.
The results indicate that certain rules remain unresolved, even for state-of-the-art LLMs.

Our survey results indicate that different moderators have different perceptions of what constitutes an adequate model: for the evaluation questions, participants exhibited only fair agreement (Fleiss' Kappa $0.34$). 
Mirroring policy decision-making standards in the subreddit, we use majority vote to summarize participants' evaluation of the models' performance. The results are shown in \autoref{fig:model-question} and \autoref{fig:model-comment}. 
In most cases, moderators would consider models with $>70\%$ precision and recall to be useful.   
Among the $23$ rules, both models are considered useful for three question rules and three comment rules. 
For some rules, such as \texttt{NSFW/Sensitive content}, \texttt{No plagiarism}, \texttt{No political agendas or moralising}, and \texttt{Depth}, moderators can accept low recall; for some, such as \texttt{Do not just post links or quotations}, moderators can accept low precision. Importantly, there are six rules ($26\%$ of all rules) for which neither model is considered useful.
These rules are \texttt{Homework}, \texttt{Privacy}, and \texttt{Scope} question rules and \texttt{No partial answer or ``placeholders''}, \texttt{Digression}, and \texttt{Sources} comment rules.
This emphasizes that, while current LLMs provide possibilities for supporting content moderation work, further improvement is needed to support useful moderation tools.

Most of the moderators were excited about the idea of having even imperfect assistant models to support their work. Two participants expressed willingness to accept a model with moderate levels of precision and recall, because \textit{``any help is better than no help.''} One stated \textit{``A tool flags a lot of things but isn't right too often? Well, it's still flagging things for my attention. A tool that doesn't flag a lot but is very frequently right? Great, I don't have to think about whether or not to remove what it's found.''} This response emphasizes the importance of model transparency. Model users often can adapt to a flawed model and make it useful as long as they are informed of its performance limitations.

Some moderators value precision more than recall because a model that mis-flagged many posts would make moderators' workload unnecessarily large. 
On the other hand, some moderators consider recall to be more important to catch all violating posts missed by moderators, especially for urgent or sensitive content, such as for the rule \texttt{Civility}. 
Moreover, some participants state that the need for the model varies according to the rules. One participant specified, \textit{``A tool would be useful for assessing and filtering breaches of more complex rules like plagiarism (especially Chat GPT use) or poor sourcing, but given the time needed to assess these claims it would need to be correct the overwhelming majority of the times it flagged something. For simpler rules like clutter, depth, or no jokes then I would like to see a tool which is able to flag the majority of comments which break the rules.''} The same participant also pointed out that they would not use a model for rules that are inherently subjective between moderators, such as \texttt{Basic facts}. Another participant further indicated the need for model explanations for complex rules.  

To inform future improvement on moderation assistant models, we asked moderators to cluster rules according to their needs. 
The results, as in \autoref{fig:bucket}, provide quantified insight into moderators' needs for model precision and recall.
We aggregate participants' answers by averaging their choice of cluster for each rule. 
Each time a participant assigned a rule to \textit{most important}, it is scored as two points; assignments to \textit{less important} are scored as one point. Rules for which a participant selected \textit{would not use} are not included in the score calculation but reported separately.
Among the $23$ rules, ten require both high precision and high recall ($>1.5$). 
For three rules, at least three participants claimed they would not use a model --- \texttt{Basic facts, Scope,} and \texttt{Sources}.
We hope our results can serve as a guide for future moderation assistant models to better align with moderators' needs.

\section{Discussion} \label{sec:sections/discussion} Overall, our findings suggest potential directions for future research on moderation assistants. First, there remains ample opportunity for model improvements. There are ten rules that moderators require both high precision and high recall, as in \autoref{fig:bucket}. For three of them, neither GPT-4 nor Llama-2 are considered useful by moderators: \texttt{Privacy} and \texttt{Homework} with low recall ($27\%$), and \texttt{No partial answer or ``placeholders''} with low precision ($63\%$). 
For four of the ten rules, although moderators mark one model as acceptable, there is still room for improvement in precision or recall: \texttt{Plagiarism}, \texttt{NSFW/Sensitive content}, and \texttt{Depth} with recall $40-62\%$, and \texttt{Do not just post links or quotations} with precision $69\%$. 
Besides, the rule \texttt{Digression}, which requires high precision, has low precision from both models ($58\%$ and $64\%$). 
Though some rules are distinct for \textit{r/AskHistorians} (e.g., \texttt{Homework}), many rules are also enforced in other subreddit communities. For example, among the 523 subreddits studied by \citet{Fiesler_2018}, $39\%$ have rules on off-topic content (similar to the \texttt{Digression} rule in our study) and $27\%$ have rules on NSFW content.
Hence, our findings underscore the existence of plenty of rules from online communities beyond toxicity detection that require exploration. 

Despite the models not performing as well as one might hope, volunteer moderators express significant enthusiasm and flexibility about having moderation assistants.
People are used to working with tools that may not be perfect, and they are skilled at finding ways to make the most of them.
In our findings, moderators explained several personal approaches they take in using a model with moderately low precision or with moderately low recall. However, in order to exercise these strategies, models' abilities must be transparent to enable users to adapt accordingly.

Furthermore, we observed varying requirements among users and rules. Even for the same rule, different moderators expressed distinct preferences for high precision or recall. This suggests that in situations where a trade-off between precision and recall is necessary, moderators may lean towards a model offering more user control, thus allowing for an adjustable trade-off. This way, they can customize the model to align with their preferences.
In addition, our participants also expressed some common preferences. For complex rules (e.g. \texttt{Plagiarism} and \texttt{Digression}), moderators prefer higher recall and model explanations over mere flagging of violations. Conversely, for simpler rules (e.g. \texttt{Current Event} and \texttt{Jokes and Humour}), they prefer higher precision.

Finally, our study showcases the importance of involving direct stakeholders (e.g. moderators) in the design and development loop for AI tools. They not only ease the alignment of model construction with users' needs but also provide insights model developers might otherwise miss.

\section{Conclusions} \label{sec:sections/conclusion} We identified gaps between the functionalities of previous models on automated moderation and the functions needed to address Reddit moderation rules. 
We conducted a model review with Hugging Face models to to see for how many rules an existing model can be used 
to detect violations. The findings reveal that a considerable number of rules are not covered by existing Hugging Face models. Even when a matching model exists, more than half of the rules require non-trivial adjustments of the model in order to be useful. We additionally identified four major types of necessary adjustments: domain adaptation, model scope extension, customization, and function shift.

Moreover, we evaluated two LLMs, GPT-4 and Llama-2, on an evaluation dataset gathered from \textit{r/AskHistorians}, and conducted a survey study to reveal whether the state-of-the-art general-purpose LLMs can handle various moderation rules. The results indicate there is still a significant gap --- these models exhibit either low recall or moderate to low precision for many of these rules. Also, moderators are not satisfied with either of the models' performance for six of $23$ rules. This limits the usability of these models in aiding moderators to identify violations in real-world moderation.

Finally, our moderator survey study provides insight into model-performance requirements for a moderation assistant model. We observed moderators' excitement about the model and willingness to be flexible to accommodate a flawed model. Moderators also offered constructive feedback on how they wish to use an assistant model differently for different rules. Such user-centered guidance should be useful in building improved, customized moderation tools in the future.


\section{Limitations} \label{sec:sections/Limitations} 
Our study has several limitations. We are primarily studying Reddit; other platforms may have different sets of rules. 
We are also focusing on English and text-only posts; violations in other languages or in other forms, such as images and videos, may face different challenges.
Furthermore, our survey study is limited to volunteer content moderators from the \textit{r/AskHistorians} community. While we anticipate that the insights gained from the moderation experience might have broader applicability across communities and for other moderators, the extent of generalization remains uncertain without further study.

Moreover, our study and findings are grounded in the set of Reddit moderation rules rather than the frequency of their violation. Consequently, we lack information regarding which rules moderators encounter most frequently and for which rules they want additional assistance.

In addition, our model testing is limited to one-shot and few-shot contexts, as opposed to other strategies, such as interactive teaching and Chain-of-Thoughts.

\section*{Acknowledgments}

This material is based upon work partially supported by the National Science Foundation under Grant No. \text{2229885} and No. \text{2131508}.

The authors are also grateful to all the reviewers who have provided helpful suggestions to improve this work, and thank members of the CLIP lab at the University of Maryland for the support on this project. We are grateful to all those who participated in our human study, without whom this research would not have been possible.

\bibliography{moderation}
\bibliographystyle{acl_natbib}
\newpage
\appendix
\onecolumn
\section{Evaluation of HF Matching Models} \label{sec:matching_eval}

We evaluated the HuggingFace matching models with our r/AskHistorians evaluation dataset. As indicated in \autoref{tab:annotation_ask}, there are four non-toxicity-related rules that have matching HF models: \texttt{NSFW/Sensitive content}, \texttt{Illegal artifacts}, \texttt{Basic Facts}, and \texttt{Jokes and humor}. Among these four matching models, the model for \texttt{Illegal artifacts} is not compatible with the HuggingFace model pipeline and lacks the appropriate code for inference. Similarly, the model for \texttt{Basic Facts} uses a model structure that is not defined within Huggingface module. 

Consequently, we evaluated only the models for \texttt{NSFW/Sensitive content} and the model for \texttt{Jokes and humor} using our dataset. The performance of the \texttt{NSFW/Sensitive content} model is: precision $57.1\%$ and recall $92.3\%$. The performance of the \texttt{Jokes and humor} model is: precision $55.6\%$ and recall $45.5\%$.

\section{Prompt for Testing LLMs}\label{sec:prompt}
\subsection{Prompt for Testing Llama-2 with the Few-shot Setting}
\texttt{<s>[INST] <<SYS>>}\\
\texttt{You are a moderator assistant. There is a moderation policy named \{rule\} that says \{rule description\}. Does the following content violate the \{rule\} policy? Simply give a yes or no answer.}\\
\texttt{<</SYS>>}\\

\noindent \texttt{\{violating example content\}}\\
\texttt{Does the following content violate the \{rule\} policy? Simply answer yes or no.[/INST] {Yes} </s><s>[INST]}\\

\noindent \texttt{\{not violating example content\}}\\
\texttt{Does the following content violate the \{rule\} policy? Simply answer yes or no.[/INST] {No} </s><s>[INST]}\\

\noindent \texttt{\{test content\}}\\
\texttt{Does the following content violate the \{rule\} policy? Simply answer yes or no.[/INST]}

\subsection{Prompt for Testing GPT-44 with the Zero-shot Setting}
\texttt{{"role": "system", "content": "You are a moderator assistant."},\\
                {"role": "user", "content": "There is a moderation policy named \{rule\} that says \{rule description\}. Does the following content violate the \{rule\} policy? Simply give a yes or no answer."},\\
                {"role": "user", "content": \{test content\}},}

\section{LLMs Performances under Zero-shot and Few-shot Settings}
\newpage
\begin{figure}[h]
    \centering
\includegraphics[width=0.7\columnwidth]{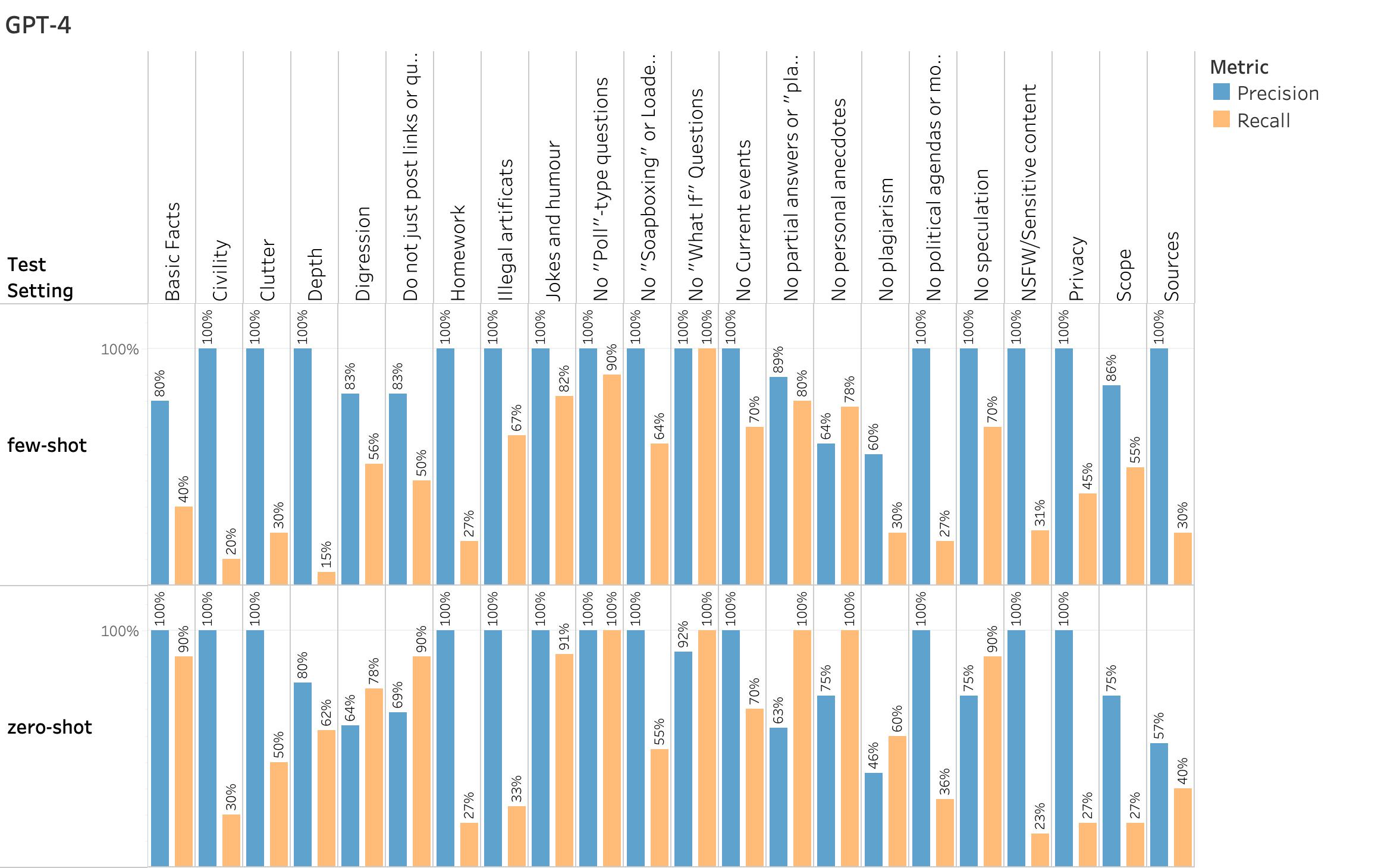}
    \caption{Model performance on detecting violations of moderation rules with GPT-4 model under few-shot setting (top) and zero-shot setting (bottom). The x-axis is the moderation rules. Each bar pair is the precision (left) and recall (right) scores on the specific rule. }
    \label{fig:gpt-append}
\end{figure}

\begin{figure}[h]
    \centering
\includegraphics[width=0.7\columnwidth]{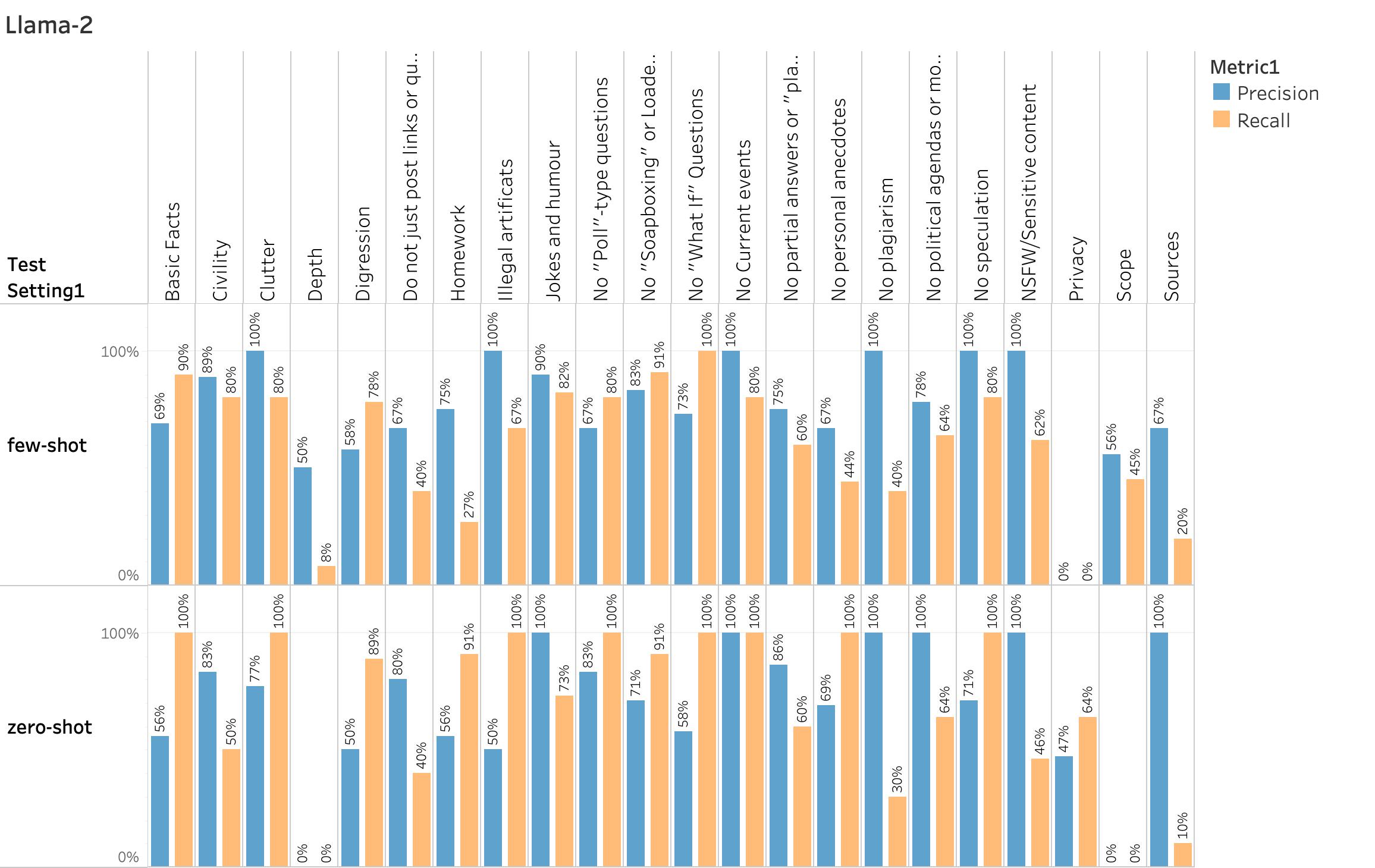}
    \caption{Model performance on detecting violations of moderation rules with Llama-2 model under few-shot setting (top) and zero-shot setting (bottom). The x-axis is the moderation rules. Each bar pair is the precision (left) and recall (right) scores on the specific rule.}
    \label{fig:llama-append}
\end{figure}

\newpage
\section{User Survey}
\begin{figure}[h]
    \centering
\includegraphics[width=0.8\columnwidth]{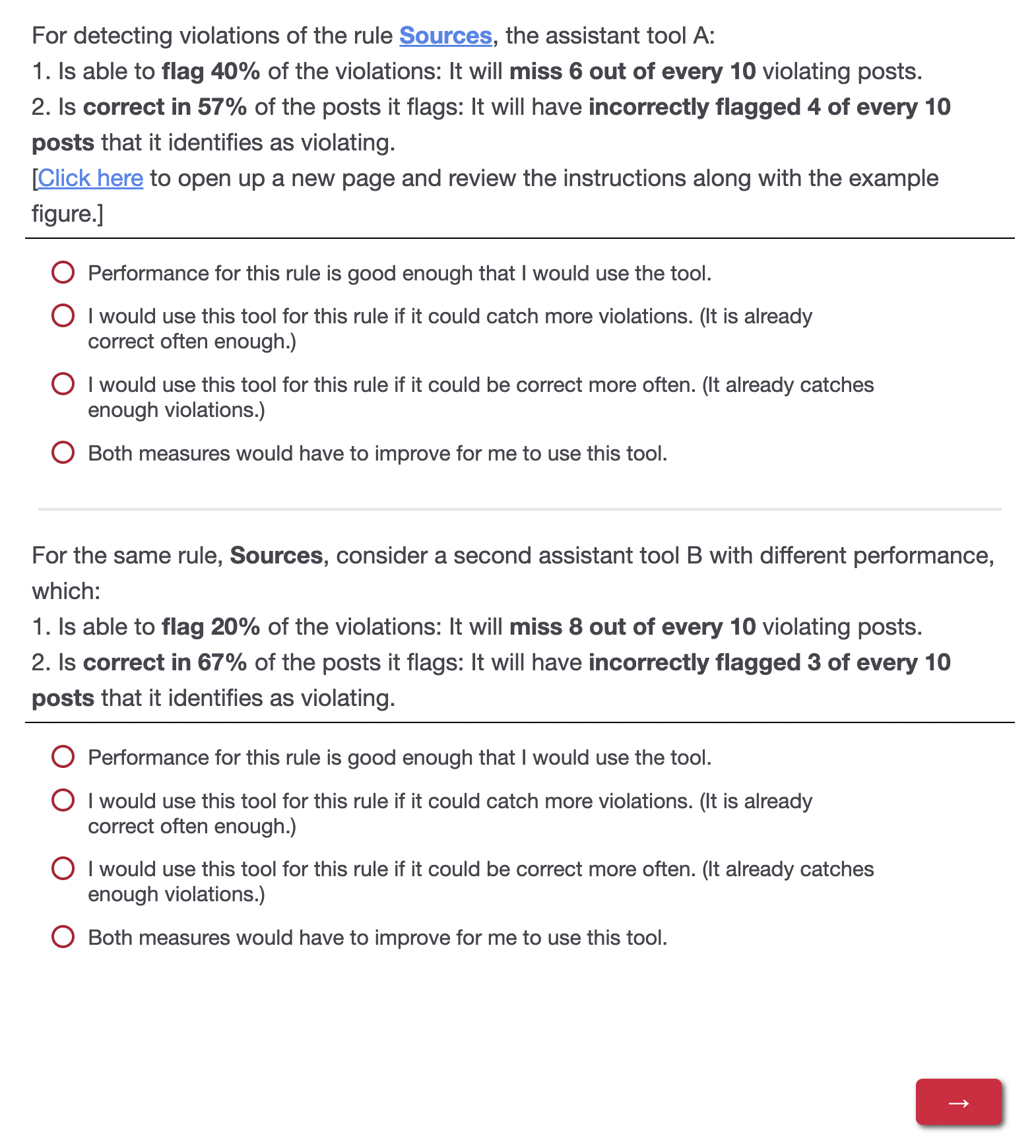}
    \caption{The evaluation survey question for the rule \texttt{Sources}. The question content is similar for all the rules.}
    \label{fig:survey-rate}
\end{figure}

\begin{figure}[t]
    \centering
\includegraphics[width=\columnwidth]{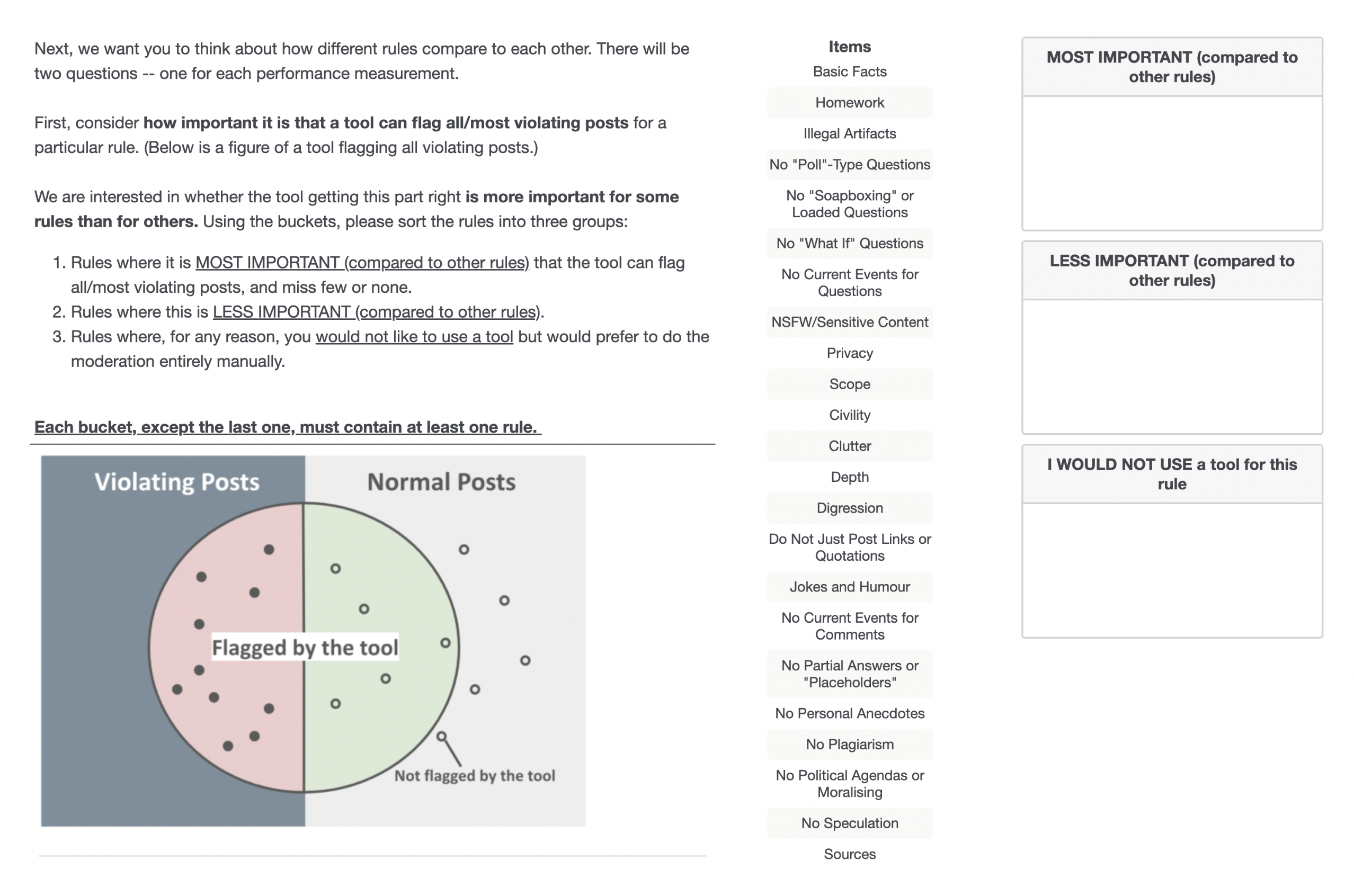}
    \caption{The clustering survey question for recall importance. The same question is asked for precision importance.  }
    \label{fig:survey-bucket}
\end{figure}

\begin{table*}[h]
  \centering
  \footnotesize
  \begin{tabular}{|>{\columncolor{gray!20}}m{3cm}|m{3.5cm}|m{3.5cm}|m{3.5cm}|}
    \hline
    \textbf{Categories of Rules} & \textbf{AskHistorian} & \textbf{Atheism} & \textbf{Movies} \\
    \hline
    \hline
    Advertising \& Commercialization & & & No Spam \& Self-promotion - Same Source Posts \\
    \hline
    \st{Consequences/ Moderation/ Enforcement} &  & & \\
    \hline
    Content/Behavior & Homework; No "what if" questions Basic facts Depth; No speculation; Jokes and humor & Proselytizing & No Antagonism/ Flamewars/ Attention Whoring - No Racial/ Sexist/ Dogwhisteling/ Homophobic Slurs; No Extraneous Comic Book Movie Submission" \\
    \hline
    Copyright & No plagiarism & & \\
    \hline
    Doxxing/Personal Info & Privacy; No personal anecdotes & Personal Attacks or Flaming & \\
    \hline
    Format & No partial answers or "placeholders"; Do not just post links or quotations & & No Ambiguous/ Misleading/ Inaccurate Information or Clickbait in The Submission Title \\
    \hline
    Harassment & Civility& Harassment or Bigotry & No Antagonism/ Flamewars/ Attention Whoring - No Racial/ Sexist/ Dogwhisteling/ Homophobic Slurs \\
    \hline
    Hate Speech & Civility & & No Antagonism/ Flamewars/ Attention Whoring - No Racial/ Sexist/ Dogwhisteling/ Homophobic Slurs \\
    \hline
    \st{Images} & & & \\
    \hline
    Links \& Outside Content & Sources & & Prohibited Popular Websites \\
    \hline
    Low-Quality Content & Scope; Clutter & Low-Effort Posts & \\
    \hline
    NSFW & NSFW/Sensitive content & & \\
    \hline
    Off-topic & Digression & Off Topic & No Extraneous Comic Book Movie Submission  \\
    \hline
    Personal Army & & Brigading & No Subreddit Brigading \\
    \hline
    Personality & Civility & Personal Attacks or Flaming & No Antagonism/ Flamewars/ Attention Whoring - No Racial/ Sexist/ Dogwhisteling/ Homophobic Slurs \\
    \hline
    Politics & No political agendas or moralising; No "Soapboxing" or loaded questions &  & \\
    \hline
    \st{Prescriptive} & & & \\
    \hline
    \st{Reddiquette} & & & \\
    \hline
    Reposting & & & No Repost or Discussion Threads of New Releases \\
    \hline
    \st{Restrictive} & & & \\
    \hline
    Spam & & Spam & No Spam \& Self-promotion - Same Source Posts \\
    \hline
    Spoilers & & & No Repost or Discussion Threads of New Releases; No Spoilers \\
    \hline
    Trolling & & No Trolling & No Ambiguous/ Misleading/ Inaccurate Information or Clickbait in The Submission Title \\
    \hline
    Voting & "Poll-type" question & & \\
    \hline
    Others & Current event; Illegal artifacts & Don't complain about the use of AAVE or slang & No Encouraging Piracy \\
    \hline
  \end{tabular}
  \caption{Subreddit rules from AskHistorian, Atheism, and Movies subreddits and their cross-referencing with categories of rule summarized by \citet{Fiesler_2018}. Rules within one category are separated by semicolons. Categories crossed out are rules that are not text-relevant or not about rule content. }\label{tab:rule_types}
\end{table*}

\begin{table*}[h]
  \centering
  \footnotesize

  \begin{tabular}{|p{2.5cm}|p{11.5cm}|}
    \hline
    \textbf{Rule} & \textbf{Rule Description} \\
    \hline
    Civility & All users are expected to behave with courtesy and politeness at all times. We will not tolerate racism, sexism, or any other forms of bigotry. This includes Holocaust denialism. Nor will we accept personal insults of any kind, and do not allow minor nitpicking of grammar or spelling. \\
    \hline
    Scope & Submissions must be about a question about the human past, a META post about the state of the subreddit, or an AMA ("Ask Me Anything") with a historical expert or panel of experts. \\
    \hline
    No Current events & To discourage off-topic discussions of current events, questions, answers, and all other comments must be confined to events that happened 20 years ago or more, inclusively (e.g., 2003 and older). \\
    \hline
    Homework & Our users aren't here to do your homework for you, but they might be willing to help. Don't just give us your essay/assignment topic and ask us for ideas. \\
    \hline
    NSFW/Sensitive content & Questions with NSFW titles will be deleted, and we will ask you to repost it with a different title. \\
    \hline
    No "Poll"-type questions & "Poll"-type questions aren't appropriate here: "Who was the most influential person in history?" or "Who was the worst general in your period?" or "Who are your Top 10 favorite people in history?" If your question includes the words "most" or "least", or "best" or "worst" (or can be reworded to include these words), it's probably a "poll"-type question. \\
    \hline
    No "Soapboxing" or Loaded Questions & All questions must allow a back-and-forth dialogue based on the desire to gain further information and not be predicated on a false and loaded premise in order to push an agenda. Example: Good Question: "People say that Nixon is the worst President of all time. Why is this so?" Bad Question: "Nixon was the worst President of all time. Why isn't Obama considered the worst?" The bad question is a fishing expedition to try to start a debate about Obama's presidency. Most of these questions will break our 20-year rule, or try to set up a debate about an issue using a long wall of text in the main post. \\
    \hline
    No "What If" Questions & Questions should be about what did happen, not what could have happened. \\
    \hline
    Privacy & We do not allow questions that pose possible privacy issues for living or recently deceased persons who are not in the public eye. The cut-off for "recent" is 100 years. \\
    \hline
    Illegal artifacts & It is our policy to disallow posts asking for further information on artifacts where there is a likelihood that the acquisition or possession of the item might be illegal, unethical, and/or run contrary to sound, historical practices. \\
    \hline
    Basic Facts & Questions looking for specific, basic facts - for the purpose of this rule, seeking a name, a date or time, a number, a location, the origin of a word, the first or last known instance/example of an object/phenomenon/etc., or a simple list of examples or facts - are not allowed as standalone threads. \\
    \hline
    Depth & An in-depth answer provides the necessary context and complexity that the given topic calls for, going beyond a simple cursory overview. Your answer should be giving context to the events being discussed, not simply listing some related facts. \\
    \hline
    Sources & We do not require sources to be preemptively listed in an answer here, but do expect that respondents be familiar with relevant and reliable literature on the topic, and that answers reflect current academic understanding or debates on the subject at hand. But sole reliance on tertiary sources for context and analysis is not allowed, and will result in the removal of a response. \\
    \hline
    No personal anecdotes & Personal anecdotes are not acceptable answers in this subreddit. \\
    \hline
    No speculation & Suppositions and personal opinions are not a suitable basis for an answer here. Warning phrases for speculation include: "I guess..." or "My guess is...", "I believe...", "I think...", "... to my understanding", "It makes sense to me that...", "It's only common sense." \\
    \hline
    No partial answers or "placeholders" & An answer should be full and complete in and of itself. \\
    \hline
    No political agendas or moralizing & This subreddit is a place for learning and open-minded discussion. As such, answers should not be written in the interests of advancing a personal agenda, but should represent a sincere effort to make an argument from the historical record. They should be constructed in keeping with the principles of the historical method - that is to say, your evidence should not be chosen selectively to support an argument that you feel is right; your argument should instead demonstrably flow from your critical engagement with an appropriate range of evidence. \\
    \hline
    Do not just post links or quotations & Do not just post links to other sites as an answer. This is not helpful. The expectation is that a user is posting to this subreddit because they are looking for the type of answer dictated by the rules in place here. Please take some time to put the links in context for the person asking the question. Avoid only recommending a source – whether that's another site, a book, or large slabs of copy-pasted text. If you want to recommend a source, please provide at least a small summary of what the source says. \\
    \hline
    No plagiarism & We have a zero-tolerance policy on blatant plagiarism, such as directly copying and pasting another person's words and trying to pass them off as your own. \\
    \hline
    Jokes and humor & A post should not consist only of a joke, a humorous remark, or a flippant comment. You can certainly include humor as part of a full and comprehensive post, but your post should not be made solely for the purpose of being funny. \\
    \hline
    Digression & All comments should be related to the topic as outlined in the original post. \\
    \hline
    Clutter & Please refrain from commenting for the sake of commenting. This includes, but is not limited to, statements about how interesting the question is, how you would like to see an answer, to "remember to come back later," to share a story that the question reminds you of, and so on. \\
    \hline
  \end{tabular}
  \caption{AskHistorian Rules and Rule Descriptions}\label{tab:rules_ask}
\end{table*}

\begin{table*}[h]
    \centering
    \footnotesize
    \begin{tabular}{|>{\columncolor{gray!20}}m{1.7cm}|m{1.7cm}|m{1.5cm}|m{0.85cm}|m{0.85cm}|m{1cm}|m{1.5cm}|m{3cm}|}
        \hline
        \textbf{Rules} & \textbf{Keywords} & \textbf{Annotated model} & \Qrule & \Qhate & \Qmatch & \Qpurpose & \Qgap \\
        \hline
        \hline
        No Trolling & trolling; provocative & \href{https://huggingface.co/socialmediaie/TRAC2020_IBEN_A_bert-base-multilingual-uncased}{TRAC2020 \_IBEN\_A\_ bert-base-multilingual-uncased} & No & Yes & Yes & & \\
        \hline
        Personal Attacks or Flaming & doxxing; cyberbullying & \href{https://huggingface.co/socialmediaie/TRAC2020_IBEN_A_bert-base-multilingual-uncased}{TRAC2020 \_IBEN\_A\_ bert-base-multilingual-uncased} & No & Yes & Yes & & \\
        \hline
        Off Topic & topic classifier & \href{https://huggingface.co/cardiffnlp/tweet-topic-21-multi}{tweet-topic-21-multi} & No & No & Yes, but with changes & Twitter topic classification & Need to be finetuned on this topic \\
        \hline
        \st{Image not submitted correctly} & & & & & & & \\
        \hline
        Spam & spam & \href{https://huggingface.co/mrm8488/bert-tiny-finetuned-sms-spam-detection}{bert-tiny-finetuned-sms-spam-detection} & No & No & Yes, but with changes & Detect message spam & 1. Need to be adapted to reddit texts rather than message texts. 2. Need to be adapted to detect self-promotion of some youtube account/website etc. \\
        \hline
        Low-Effort Posts & low effort; low quality & NA & No & No & No& & \\
        \hline
        Proselytizing & proselytizing; Soapbox & NA & No & No & No&  & \\
        \hline
        Harassment or Bigotry & harassment; bigotry & \href{https://huggingface.co/KoalaAI/Text-Moderation}{Text-Moderation} & No & Yes & Yes, but with changes & & any and all curse words are permitted \\
        \hline
        Brigading & brigading & NA - Toxicity & No & Yes & Yes, but with changes & & These models may be able to detect some of the brigading comments, but there is no specific model for brigading and some brigading encouraging comments may not be caught by the toxicity models \\
        \hline
    \end{tabular}
    \caption{Model review for the Atheism subreddit. The second and third columns are the keywords used for model searching and the best-matching model from hugging face. The rest columns are annotation results.  Crossed-out rules are rules not related to text input. }\label{tab:annotation_atheism}
\end{table*}

\clearpage
\footnotesize
\begin{longtable}[h]{|>{\columncolor{gray!20}}m{1.7cm}|m{1.7cm}|m{1.5cm}|m{0.85cm}|m{0.85cm}|m{1cm}|m{1.5cm}|m{3cm}|}
  \hline
  
  \textbf{Rules} & \textbf{Keywords} & \textbf{Annotated model} & \textbf{\Qrule} & \textbf{\Qhate} & \textbf{\Qmatch} & \textbf{\Qpurpose} & \textbf{\Qgap} \\
  \hline
  \endfirsthead

  \multicolumn{8}{c}%
  {\tablename\ \thetable\ -- \textit{Continued from previous page}} \\
  \hline
  \textbf{Rules} & \textbf{Keywords} & \textbf{Annotated model} & \textbf{\Qrule} & \textbf{\Qhate} & \textbf{\Qmatch} & \textbf{\Qpurpose} & \textbf{\Qgap} \\
  \hline
  \endhead
        
        \st{Do Familiarize Yourself With Our Rules} & & & & & & & \\
        \hline
        Incivility - No Antagonism/ Flamewars/ Attention Whoring - No Racial/ Sexist/ Dogwhistling/ Homophobic Slurs & offensive; hatespeech & \href{https://huggingface.co/KoalaAI/Text-Moderation}{Text-Moderation} & No & Yes & Yes, but with changes & Yes & Quoting something offensive from a movie is okay but needs to be put in quotation marks. \\
        \hline
        No Spam \& Self-promotion - Same Source Posts & spam & \href{https://huggingface.co/mrm8488/bert-tiny-finetuned-sms-spam-detection}{bert-tiny-finetuned-sms-spam-detection} & No & No & Yes, but with changes & No & Detect message spam - 1. Need to be adapted to Reddit texts rather than message texts. 2. Need to be adapted to detect self-promotion of some YouTube account/website etc. \\
        \hline
        \st{No Image Posts \& Memes} & & & & & & & \\
        \hline
        No Ambiguous/ Misleading/ Inaccurate Information or Clickbait in The Submission Title - misinformation & misinformation & \href{https://huggingface.co/roupenminassian/TwHIN-BERT-Misinformation-Classifier}{TwHIN-BERT-Misinformation-Classifier} & No & No & Yes, but with changes & No & Detect misinformation for Twitter posts - 1. Need to be adapted to title-like texts. 2. Need to be adapted to a movie-related context \\
        \hline
        No Ambiguous/ Misleading/ Inaccurate Information or Clickbait in The Submission Title - Clickbait & clickbait & \href{https://huggingface.co/valurank/distilroberta-clickbait}{distilroberta-clickbait} & No & No & Yes & Yes & \\
        \hline
        No Extraneous Comic Book Movie Submission & misinformation & \href{https://huggingface.co/roupenminassian/TwHIN-BERT-Misinformation-Classifier}{TwHIN-BERT-Misinformation-Classifier} & No & No & Yes, but with changes & No & Detect misinformation for Twitter posts - Need to be adapted to a movie-related context \\
        \hline
        No Repost & rule-based & & Yes & No & & & \\
        \hline
        No Discussion Threads of New Releases & rule-based & & Yes & No & & & \\
        \hline
        No Spoilers & spoiler & \href{https://huggingface.co/bhavyagiri/roberta-base-finetuned-imdb-spoilers}{roberta-base-finetuned-imdb-spoilers} & No & No & Yes & Yes & \\
        \hline
        No Encouraging Piracy & piracy & \href{https://huggingface.co/HannahDK/Classification}{Classification} & No & No & Yes, but with changes & No & Detect if a website is a piracy website - Need to detect not only piracy website but also comments that encourage piracy \\
        \hline
        No Subreddit Brigading & brigading & NA - Toxicity & No & Yes & Yes, but with changes & No & These models may be able to detect some of the brigading comments, but there is no specific model for brigading and some brigading encouraging comments may not be caught by the toxicity models \\
        \hline
        Prohibited Popular Websites & rule-based & & Yes & No & & & \\
        \hline
    \caption{Model review for the Movies subreddit. The second and third columns are the keywords used for model searching and the best-matching model from hugging face. The rest columns are annotation results.  Crossed-out rules are rules not related to text input. }\label{tab:annotation_movies}
\end{longtable}

\clearpage
\footnotesize
\begin{longtable}[h]{|>{\columncolor{gray!20}}m{1.7cm}|m{1.7cm}|m{1.5cm}|m{0.85cm}|m{0.85cm}|m{1cm}|m{1.5cm}|m{3cm}|}
  \hline
  
  \textbf{Rules} & \textbf{Keywords} & \textbf{Annotated model} & \textbf{\Qrule} & \textbf{\Qhate} & \textbf{\Qmatch} & \textbf{\Qpurpose} & \textbf{\Qgap} \\
  \hline
  \endfirsthead

  \multicolumn{8}{c}%
  {\tablename\ \thetable\ -- \textit{Continued from previous page}} \\
  \hline
  \textbf{Rules} & \textbf{Keywords} & \textbf{Annotated model} & \textbf{\Qrule} & \textbf{\Qhate} & \textbf{\Qmatch} & \textbf{\Qpurpose} & \textbf{\Qgap} \\
  \hline
  \endhead

  Civility & toxicity; hatespeech & \href{https://huggingface.co/KoalaAI/Text-Moderation}{Text-Moderation} & No & Yes & Yes & & The model may not be able to catch Holocaust denialism, which is vital in this case. \\
  \hline
  Current event & time/year prediction for event; history event; event extractor & \href{https://huggingface.co/HiTZ/GoLLIE-7B}{GoLLIE-7B} & No & No & Yes, but with changes & Information extraction & 1. The model needs to be adapted to question-contexts and extract event names from the question. 2. Then by matching the event name with Wikipedia page, we may get the year or time of the event. \\
  \hline
  Homework & homework & NA & No & No & No & & \\
  \hline
  Scope & topic classifier & \href{https://huggingface.co/cardiffnlp/tweet-topic-21-multi}{tweet-topic-21-multi} & No & No & Yes, but with changes & Twitter topic classification & Need to be fine-tuned to this topic. \\
  \hline
  NSFW/Sensitive content & NSFW; sexual & \href{https://huggingface.co/michellejieli/NSFW_text_classifier}{NSFW\_text \_classifier} & No & No & Yes & & \\
  \hline
  "Poll-type" question & poll & NA & No & No & No & & \\
  \hline
  "Soapboxing" or loaded questions & soapboxing; agenda pushing & NA & No & No & No & & \\
  \hline
  No "what if" questions & hypothetical; counterfactual; alternative history & NA & No & No & No & & \\
  \hline
  Privacy & PII; personal private & \href{https://huggingface.co/FelipeCasali-USP/lgpd_pii_identifier}{lgpd\_pii\_ identifier} & No & No & Yes, but with changes & Identify sensitive data in the scope of LGPD. The goal is to have a tool to identify document numbers like CNPJ, CPF, people's names, and other kinds of sensitive data, allowing companies to find and anonymize data according to their business needs, and governance rules. & Need to be adapted to the sensitive data here \\
  \hline
  Illegal artifacts & illegal; safety & \href{https://huggingface.co/PKU-Alignment/beaver-dam-7b}{beaver-dam-7b} & No & No & Yes & & \\
  \hline
  Basic facts & basic facts; factoid & \href{https://huggingface.co/Lurunchik/nf-cats}{nf-cats} & No & No & Yes & & \\
  \hline
  Depth & depth & NA & No & No & No & & \\
  \hline
  Sources & reliable source & NA & No & No & No & & \\
  \hline
  No personal anecdotes & personal anecdote & \href{https://huggingface.co/Reggie/muppet-roberta-base-joke_detector}{muppet-roberta-base-joke\_detector} & No & No & Yes, but with changes & Detect jokes, stories, and anecdotes & The model is trained with 2000 jokes. Here the detection target is not about jokes. In addition, here the need is focusing on personal anecdotes. \\
  \hline
  No speculation & speculation; factual; fact-check & \href{https://huggingface.co/saattrupdan/verdict-classifier-en}{verdict-classifier-en} & No & No & Yes, but with changes & Fact-checking model for detecting misinformation & The model does not detect speculation or personal opinion. Here the detection target is the speculation of history events, which can be especially hard for such a fact-checking model. \\
  \hline
  No partial answers or "placeholders" & partial answer & NA & No & No & No & & \\
  \hline
  No political agendas or moralizing & political agenda; moralizing & NA & No & No & No & & \\
  \hline
  Do not just post links or quotations & rule-based & NA & Yes & No & & & \\
  \hline
  No plagiarism & plagiarism & \href{https://huggingface.co/jpwahle/longformer-base-plagiarism-detection}{longformer-base-plagiarism-detection} & No & No & Yes, but with changes & Detect machine-paraphrased plagiarisms & The real-world source of plagiarism is way larger than the training and testing dataset size of this model. \\
  \hline
  Jokes and humor & joke; humor; flippant & \href{https://huggingface.co/Reggie/muppet-roberta-base-joke_detector}{muppet-roberta-base-joke\_detector} & No & No & Yes & & \\
  \hline
  Digression & relevance & \href{https://huggingface.co/tinkoff-ai/response-quality-classifier-large}{response-quality-classifier-large} & No & No & Yes, but with changes & Rate the relevancy and specificity of a response within a dialogue & Need to be adapted to the Reddit (question-comment) texts. \\
  \hline
  Clutter & clutter & NA & No & No & No& & \\
  \hline

\caption{Model review for the AskHistorian subreddit. The second and third columns are the keywords used for model searching and the best-matching model from Hugging Face. The rest of the columns are annotation results. Crossed-out rules are rules not related to text input.}
\label{tab:annotation_ask}
\end{longtable}

\label{sec:appendix}


\end{document}